\documentclass[lettersize,journal]{IEEEtran}

\usepackage{algorithmic}
\usepackage{array}
\usepackage[caption=false,font=normalsize,labelfont=sf,textfont=sf]{subfig}
\usepackage{textcomp}
\usepackage{stfloats}
\usepackage{url}
\usepackage{verbatim}
\def\BibTeX{{\rm B\kern-.05em{\sc i\kern-.025em b}\kern-.08em
    T\kern-.1667em\lower.7ex\hbox{E}\kern-.125emX}}
\usepackage{balance}

\usepackage{graphics} 
\usepackage{epsfig} 
\usepackage{times} 
\usepackage{amsmath} 
\usepackage{amssymb}  
\usepackage{amsfonts,amssymb}
\usepackage{bm}
\usepackage{graphicx}
\usepackage{float} 
\usepackage[outdir=./]{epstopdf}
\usepackage{color}
\usepackage[dvipsnames]{xcolor}
\usepackage{setspace}
\usepackage{multirow}
\usepackage{booktabs}

\makeatletter
\renewcommand{\maketag@@@}[1]{\hbox{\m@th\normalsize\normalfont#1}}%
\makeatother

\title{Human-Like Robot Impedance Regulation Skill Learning from Human-Human Demonstrations
}

\author{Chenzui Li, Xi Wu, Yiming Chen, Tao Teng, Xuefeng Zhang, Sylvain Calinon, \IEEEmembership{Member, IEEE}, \\ Darwin Caldwell, \IEEEmembership{Fellow, IEEE}, and Fei Chen, \IEEEmembership{Senior Member, IEEE}

\thanks{This work was supported in part by the Research Grants Council of the Hong Kong SAR under Grant 24209021, 14222722, 14211723 and C7100-22GF, in part by the Ningbo “Innovation Yongjiang 2035” Key R$\&$D Programme under Grant 2024Z202, and in part by the InnoHK of the Government of Hong Kong via the Hong Kong Centre for Logistics Robotics. (\textit{Corresponding author: Fei Chen)}}
\thanks{Chenzui Li, Xi Wu, Yiming Chen, Tao Teng, and Fei Chen are with the Collaborative and Versatile Robots (CLOVER) Laboratory, Department of Mechanical and Automation Engineering, T-Stone Robotics Institute, The Chinese University of Hong Kong, Hong Kong (e-mail: {czli@mae.cuhk.edu.hk, xwu@mae.cuhk.edu.hk, ymchen@mae.cuhk.edu.hk, tao.teng@cuhk.edu.hk, f.chen@ieee.org}).}
\thanks{Xuefeng Zhang is with the College of Science and Technology, Ningbo University, Ningbo, China (e-mail: zhangxuefeng@nbu.edu.cn).}
\thanks{Sylvain Calinon is with the Idiap Research Institute, Martigny, Switzerland (e-mail: {sylvain.calinon@idiap.ch}).}
\thanks{Darwin Caldwell is with the Department of Advanced Robotics, Istituto Italiano di Tecnologia, 16163 Genoa, Italy (e-mail: {darwin.caldwell@iit.it}).}

}

\begin{document}

\maketitle

\begin{abstract}
Humans are experts in physical collaboration by leveraging cognitive abilities such as perception, reasoning, and decision-making to regulate compliance behaviors based on their partners' states and task requirements. Equipping robots with similar cognitive-inspired collaboration skills can significantly enhance the efficiency and adaptability of human-robot collaboration (HRC). This paper introduces an innovative Human-Inspired Impedance Regulation Skill Learning framework (HI-ImpRSL) for robotic systems to achieve leader-follower and mutual adaptation in multiple physical collaborative tasks. The proposed framework enables the robot to adapt its compliance based on human states and reference trajectories derived from human-human demonstrations. By integrating electromyography (EMG) signals and motion data, we extract endpoint impedance profiles and reference trajectories to construct a joint representation via imitation learning. An LSTM-based module then learns task-oriented impedance regulation policies, which are implemented through a whole-body impedance controller for online impedance adaptation. Experimental validation was conducted through collaborative transportation, two interactive Tai Chi pushing hands, and collaborative sawing tasks with multiple human subjects, demonstrating the ability of our framework to achieve human-like collaboration skills and the superior performance from the perspective of interactive forces compared to four other related methods.
\end{abstract}


\section{Introduction} \label{introduction}
\begin{figure}[!tbp]
\centering
\includegraphics[width=0.9\linewidth]{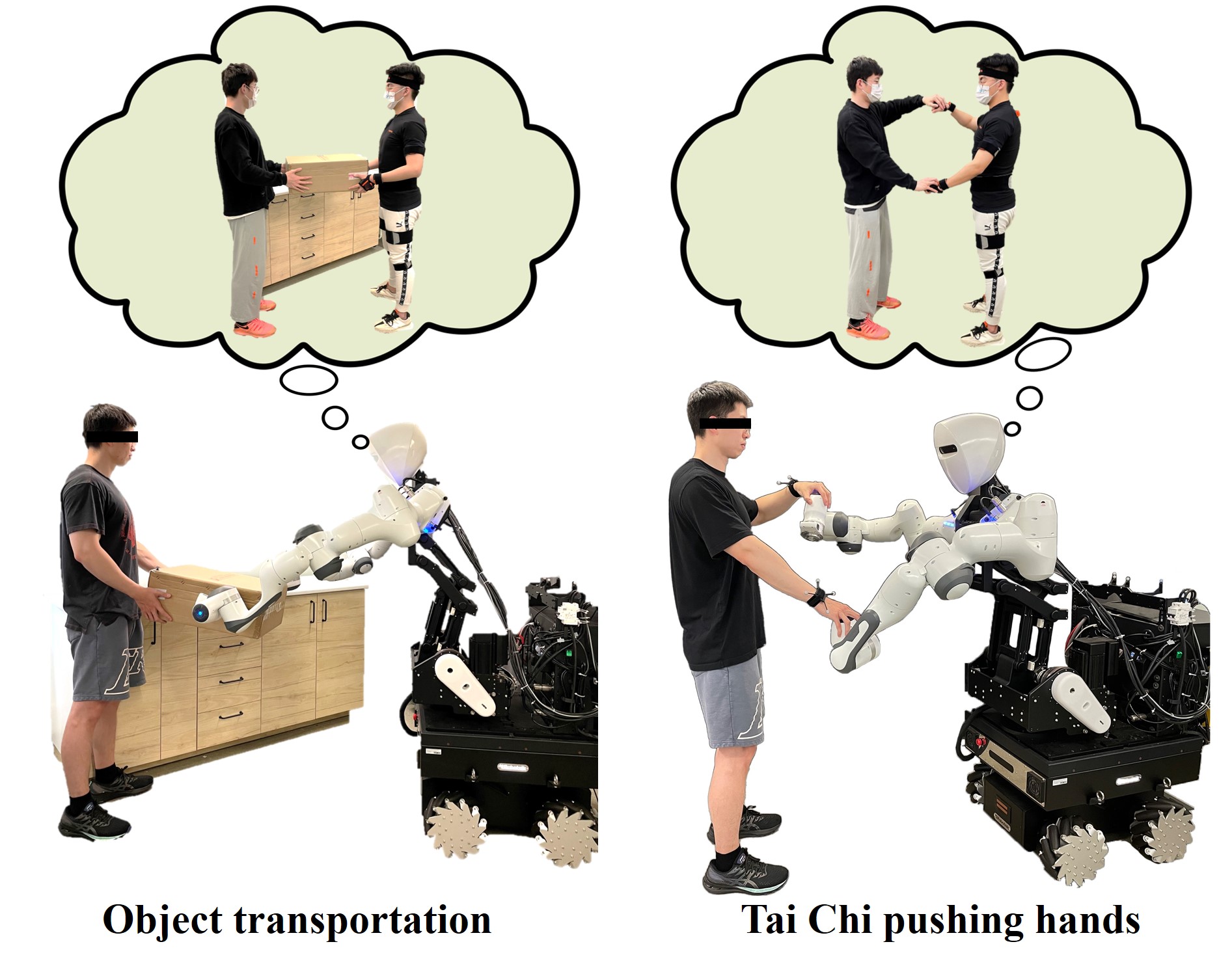} 
\caption{Illustration of human-robot physical collaborative tasks. In these scenarios, the cobot should be able to perform different compliance behaviors accordingly because physical interaction exists between the human and the cobot. We expect to learn from the human-human demonstrations and apply them to a human-like cobot.}
\label{Fig.tai chi} \vspace{-3mm}
\end{figure}

\IEEEPARstart{C}{ollaborative} robots (cobots) have emerged as a practical route to more intuitive and efficient human-robot collaboration (HRC) across industrial and domestic settings. By combining human adaptability with robotic precision, collaborative manipulation improves task flexibility and effectiveness while alleviating human workload and fatigue-related errors \cite{sheng2025human}. Consider the various collaborative tasks in Fig. \ref{Fig.tai chi}, distinct collaboration skills are required for cobots, which operate at the cognitive level and are embodied physically through impedance regulation skills. For instance, during human-robot transportation of bulky objects, humans take on the role of leaders for multiple operations while cobots undertake physical work and regulate impedance based on the human compliance behaviors, reflecting cognitive processes such as understanding leader intentions and dynamically adapting to varying compliance \cite{nemec2016bimanual}. Tai Chi pushing hands tasks require cobots to produce inverse compliance behaviors compared to the human partner in different stages, highlighting not only the need for precise physical actions but also the cognitive ability to recognize partner strategies and adapt to changing interaction dynamics. Nevertheless, it remains challenging for cobots to swiftly acquire these impedance regulation skills through traditional model-based methods and effectively apply them in human-robot collaboration \cite{8907351}.

Learning from Demonstration (LfD) is an effective approach that enables cobots to acquire and imitate these skills for HRC \cite{ravichandar2020recent}. Most LfD approaches involve a single human instructing a cobot through guided collaboration, commonly referred to as kinesthetic teaching to achieve tasks such as collaborative object carrying \cite{rozo2016learning} and scarf hanging \cite{7139990}. Other studies concentrate on learning from human-human demonstrations, which directly capture physical behaviors (e.g., motion, compliance) of one partner and record relevant kinematic data \cite{7989334} and dynamic data \cite{zhang2022electromyography}.

Trajectory and impedance are widely used to represent human kinematic and dynamic behaviors \cite{wu2024simultaneous}. On the one hand, Gaussian Mixture Model (GMM) and Dynamic Movement Primitives (DMP) are frequently used to represent complex human demonstration motions with a few parameters that can generate reference trajectories for cobots \cite{CalinonLee19}. The Task-Parameterized Gaussian Mixture Model (TP-GMM) represents demonstration motions from several different perspectives for integrating task information, which allows the cobots to adjust trajectories to new situations \cite{calinon2016tutorial}. On the other hand, stiffness and damping parameters are commonly applied to characterize the physical behavior of humans. The cobot impedance variables can be inferred based on the dynamic interaction model and learned Cartesian trajectories with optimization methods. Such approaches have been successfully applied to various collaborative tasks, such as object assembly \cite{jung2024touch}, object lifting \cite{yang2018dmps}, sawing \cite{cao2023passive}, and drag trajectory tracking motion in open space \cite{zhang2025intention}. Regardless, these approaches acquire data from kinesthetic demonstration and the optimized impedance variations of the cobots may not align with those encoded physical behaviors during human-human demonstrations. Other approaches utilize biosignals, such as electromyography (EMG), to directly record human muscle bioelectrical activity, thereby offering valuable insights into human impedance \cite{luo2024impedance}. An EMG-based adaptive variable impedance control has been validated in collaborative sawing and wiping \cite{wang2025adaptive}. Another framework has adapted robot trajectories with force feedback, AR motion prediction, and EMG-based impedance to reduce interaction force and maintain tracking accuracy \cite{luo2024physical}. However, these methods focus on replicating physical behaviors without considering their dependence on the partner’s state and task dynamics, which limits task generalization due to neglecting adaptive regulation. To address this, our study highlights learning impedance policies driven by cognitive principles of different collaboration skills.

A variable impedance controller (VIC) is required to reproduce the learned impedance regulation skills by modulating the desired stiffness and damping of the cobot \cite{sharifi2021impedance}. Among these methods, the human-in-the-loop VIC is crucial for improving safety and efficiency in HRC \cite{wang2023based}. For instance, a neuroadaptive impedance control with estimation of human-robot interaction (HRI), EMG-driven stiffness, prescribed performance, and neural compensation enables natural stable interaction with improved accuracy \cite{liu2025neuroadaptive}. Nevertheless, usability and human perception should be considered or learned for designing a collaborative controller \cite{maccarini2022preference}. The above issues can be solved by proposing a novel VIC based on the impedance regulation skills learned from human-human demonstrations since the regulation policies can be extracted from teaching data while usability is inherently ensured. By integrating cognitive-inspired impedance regulation with imitation learning methods, the proposed VIC framework enables robots to adapt dynamically to their human partners, contributing to the development of human-like collaboration skills learning.

In this article, we propose a novel human-inspired impedance regulation skill learning framework (HI-ImpRSL) designed to enable cobots to acquire collaboration skills based on human-human demonstrations. Diverging from prevalent methodologies, HI-ImpRSL extracts the physical behaviors exhibited by human demonstrators and assimilates the proper compliance responses based on the partner's dynamical states in collaborative tasks. Consequently, the cobot can be applied to execute a reference trajectory derived from task demonstrations and to simultaneously adjust compliance in response to the feedback of the human partner with a variable impedance controller. Three collaborative tasks, including collaborative transportation, confrontational unimanual and bimanual Tai Chi pushing hands, and collaborative sawing, are raised for validation with various subjects.

To summarize, the main contributions of this work are described as follows:
\begin{itemize}
  \item [1)]
  HI-ImpRSL framework is proposed for cobots to acquire collaborative compliance behaviors of humans. The imitation learning method is applied for joint motion-impedance representation from human demonstrations. Specifically, TP-GMM is adopted to represent demonstrated trajectory and impedance profiles, which are extendable to new situations (e.g., start-goal configurations, collaborator kinematics). The Linear Quadratic Tracking (LQT) based method is then applied for cobot motion and impedance planning, which explicitly couples compliance behaviors with the reference trajectories.
  \item [2)]
  An LSTM-based impedance regulation module is constructed to map the compliance variation during human-human demonstrations. This module can dynamically adjust the cobot regulatory factor online, using EMG feedback from the partner to scale the learned impedance profile. The generalization of this module to different subjects has been verified through both human-human and human-robot collaboration.
  \item [3)]
  A novel whole-body impedance controller has been designed for our human-like mobile cobot with a self-designed torso and two manipulators included, which can generate desired stiffness and damping during movement in Cartesian space while coordinating different components through proper torque commands. 
  \item [4)]
  A comprehensive comparison with four other methods shows the superior performance of the HI-ImpRSL by collaborating with various human subjects.
\end{itemize}

\section{Impedance Regulation Skills} \label{skills}
This section presents the relationship between impedance regulation skills and collaboration skills with two typical modes in human-human collaboration (as shown in Fig. \ref{Fig.physical collaboration skill}), and the definition of impedance regulation skills is given. An EMG-based stiffness estimation method is then introduced for human compliance representation, which is the preliminary of learning impedance regulation skills in demonstrations.

\subsection{Definition} \label{skill definition}
\begin{figure*}[htbp]
\centering
\setlength{\abovecaptionskip}{0.1cm}
\includegraphics[width=0.9\linewidth]{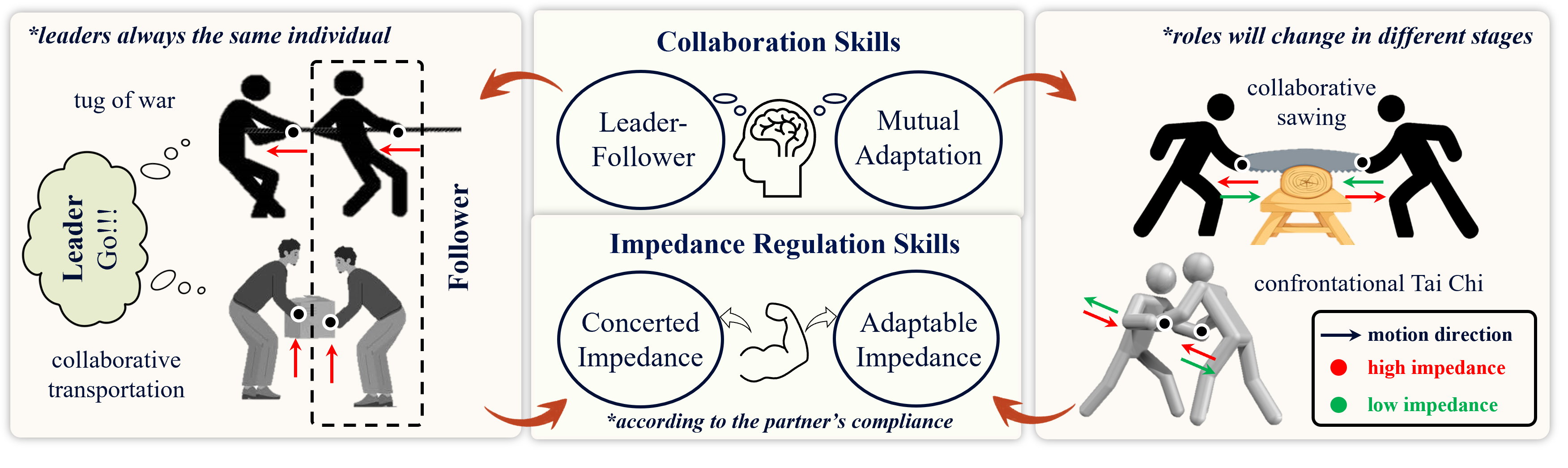} 
\caption{Schematic diagram of two typical collaboration skills and the corresponding impedance regulation skills. \textit{Leader-follower:} the follower's physical behaviors are concluded as performing concerted impedance according to the leader. \textit{Mutual adaptation:} roles are changing during these tasks and the physical behaviors of collaborators are defined as performing adaptable impedance based on the partners.}
\label{Fig.physical collaboration skill} \vspace{-3mm}
\end{figure*}

Impedance regulation skills represent a foundational aspect of collaboration skills at a low level. Collaboration skill has been defined as regulating the exchange of mechanical energy \cite{5379513} or seamless adaptation of autonomy level \cite{9501975} based on tasks and a mutual understanding of the partner's intentions. To conclude, collaboration skills involve generating appropriate behaviors based on comprehensive feedback with cognitive processes. For example, during a collaborative sawing task shown in Fig. \ref{Fig.physical collaboration skill}, one agent may demonstrate stiffly in the task by vigorously pulling the saw at one stage while exhibiting adaptability by gently pushing the saw back at another stage in response to the partner’s actions \cite{ZENG2021103668}. Consequently, collaboration skills can be conceptualized as regulatory compliance behaviors tailored to meet both task requirements and partner feedback at physical levels.

Representing human compliance behaviors can teach cobots the corresponding impedance regulation skills from human-human demonstrations since they are the most efficient capacities to transfer to the robots. Human-like impedance regulation skills have been successfully learned by human-like cobots with learning frameworks through physical human-robot interfaces \cite{6224904}. Nevertheless, these skills are mostly applied in the manipulation and teleoperation scenarios, in which the human compliance behaviors are directly transferred to the cobots whether offline or online \cite{arashijrr}. In human-human collaboration, two quintessential physical collaboration modes, leader-follower and mutual adaptation, are widely observed, and their differences serve as a comprehensive foundation for studying the regulation of compliance behaviors (Fig. \ref{Fig.physical collaboration skill}).

\textbf{Leader-follower mode}: Followers are expected to align closely with leaders from the perspective of compliance behaviors. Consequently, collaborators will demonstrate analogous endpoint impedance variations relative to their partners throughout the task execution. For example, during a collaborative transportation task, one individual leads the movement, and the other adapts their compliance to follow and support the shared goal. This mode is characterized by unilateral compliance regulation, where the follower reacts to the leader’s compliance to ensure smooth and efficient collaboration.

\textbf{Mutual adaptation mode}: Collaborators are expected to adapt their roles at different stages of the task, thereby exhibiting adaptable compliance in relation to their partners. For instance, in Tai Chi pushing hands, both participants continuously adjust their compliance and interaction forces in response to each other’s movements. This mode requires bidirectional compliance regulation and dynamic role switching, which emphasizes mutual understanding and adaptation.

These two modes highlight distinct collaboration dynamics: leader-follower tasks focus on reactive compliance, while mutual adaptation tasks require simultaneous proactive and reactive behaviors. Studying these modes provides insight into how humans regulate their compliance based on both task demands and their partner’s actions, offering a foundation for understanding physical collaboration.

To illustrate their regulatory compliance behaviors in terms of impedance, the impedance regulation skills of the human based on partner compliance can be described as:
\begin{equation}\label{human impedance relationship}
  \bm{K}_e^c(t) = h(\bm{K}_e^p(t)),
\end{equation}
where $\bm{K}_e^c(t)$ and $\bm{K}_e^p(t)$ are the contact point stiffness of the collaborator and the partner, $h$ represents the function which constructs the relationship of physical compliance along movements. The damping matrix is set as $\bm{D}_e^c(t) = \delta \sqrt{\bm{K}_e^c(t)}$ with a predefined scaling factor $\delta$.

\subsection{EMG-Based Stiffness Estimation} \label{emg-based stiffness}
\begin{figure}[!tbp]
\centering
\setlength{\abovecaptionskip}{0.cm}
\includegraphics[width=0.8\linewidth]{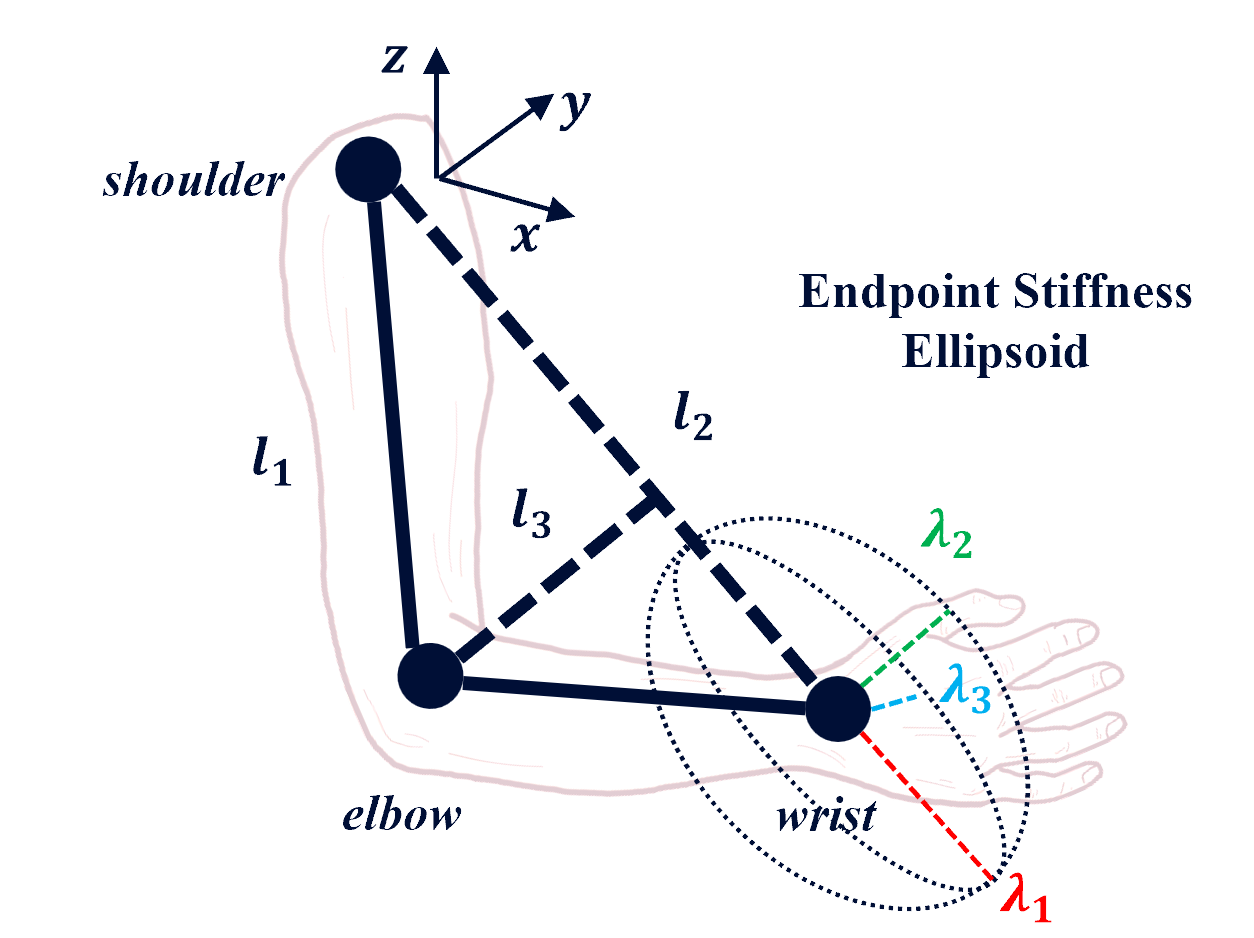} 
\caption{Definition of the endpoint stiffness ellipsoid under a human arm configuration. A simplified arm geometric configuration is constructed with the connection between the shoulder joint, elbow joint, and wrist joint. The major, median, and minor principal axes of the endpoint stiffness ellipsoid are $\lambda_1$, $\lambda_2$, and $\lambda_3$, respectively.}
\label{Fig.stiffness ellipsoid} \vspace{-4mm}
\end{figure}

Human arm endpoint stiffness should be estimated to represent compliance behaviors and further acquire impedance regulation skills during demonstrations. Stochastic perturbation techniques are one of the most reliable methods to directly construct the relationship between force and displacement and therefore identify the characteristics of the human endpoint stiffness in Cartesian space. These methods are not feasible to be directly applied in human-human demonstrations or human-robot collaboration since real-time estimation is acquired. Hence, current researchers have designed a calibration process to extract the stiffness-related biosignal features (e.g., EMG signals) for estimating arm endpoint stiffness \cite{osu1999}. Firstly, we construct the positive-definite endpoint stiffness matrix through a geometric approach based on the process proposed in \cite{s20185357}: 
\begin{equation}\label{eigendecomposition}
\bm{K}_e = \bm{V} \bm{H} \bm{V}^T,
\end{equation}
where $\bm{V}$ is an orthonormal matrix with the normalized eigenvectors of $\bm{K}_e$, and $\bm{H}$ represents a diagonal matrix composed of the eigenvalues of $\bm{K}_e$. A simplified human arm structure in Cartesian space is applied to provide the geometric information to construct these matrices. As shown in Fig. \ref{Fig.stiffness ellipsoid}, the endpoint stiffness ellipsoid can be defined under a specific arm configuration.

We define that vector $\bm{l}_1$ starts from the shoulder joint to the elbow joint and vector $\bm{l}_2$ starts from the shoulder joint to the wrist joint. $\bm{l}_3$ is a perpendicular vector starts from the elbow joint to the vector $\bm{l}_1$ while $\bm{l}_4$ represents a vector which is perpendicular to the arm triangle plane. Hence, the major, median, and minor principal axis $\lambda_1$, $\lambda_2$, and $\lambda_3$ of the endpoint stiffness ellipsoid are determined as $\bm{l}_2$, $\bm{l}_3$, and $\bm{l}_4$. The matrices $\bm{V} \in \mathbb{R}^{3 \times 3}$ and $\bm{H} \in \mathbb{R}^{3 \times 3}$ can be written as:
\begin{equation}\label{V}
\bm{V} = \left [ \begin{matrix} \
    \displaystyle{\frac{\bm{l}_2}{||\bm{l}_2||}} \ \displaystyle{\frac{(\bm{l}_2 \times \bm{l}_1) \times \bm{l}_2}{||\bm{l}_2 \times \bm{l}_1) \times \bm{l}_2||}} \ \displaystyle{\frac{\bm{l}_2 \times \bm{l}_1}{||\bm{l}_2 \times \bm{l}_1||}} \
    \end{matrix} \right ],
\end{equation}
\begin{equation}\label{H}
\bm{H} = \alpha(A) \cdot \bm{H}_s = \alpha(A) \cdot diag(1, \frac{a_1}{d_1}, a_2 d_2),
\end{equation}
where $\alpha(A)$ represents the synergistic contribution of the muscle co-contractions with $\alpha(A) = b_1 A + b_2$, $A$ denotes the comprehensive muscle activation level, $b_1$ and $b_2$ are the model parameters to be identified; $\bm{H}_s \in \mathbb{R}^{3 \times 3}$ is the diagonal matrix that also contains the personalized parameters $a_1, a_2$ and configuration dependence eigenvalues. Their values are given by the length of vector $l_2$ and $l_3$:
\begin{equation}\label{d}
d_1 = ||\bm{l}_2||, \ \ d_2 = ||\bm{l}_3|| = ||\bm{l_1} \cdot \frac{(\bm{l}_2 \times \bm{l}_1) \times \bm{l}_2}{||\bm{l}_2 \times \bm{l}_1) \times \bm{l}_2||}||.
\end{equation}

\begin{figure*}[htbp]
\centering
\includegraphics[width=0.98\linewidth]{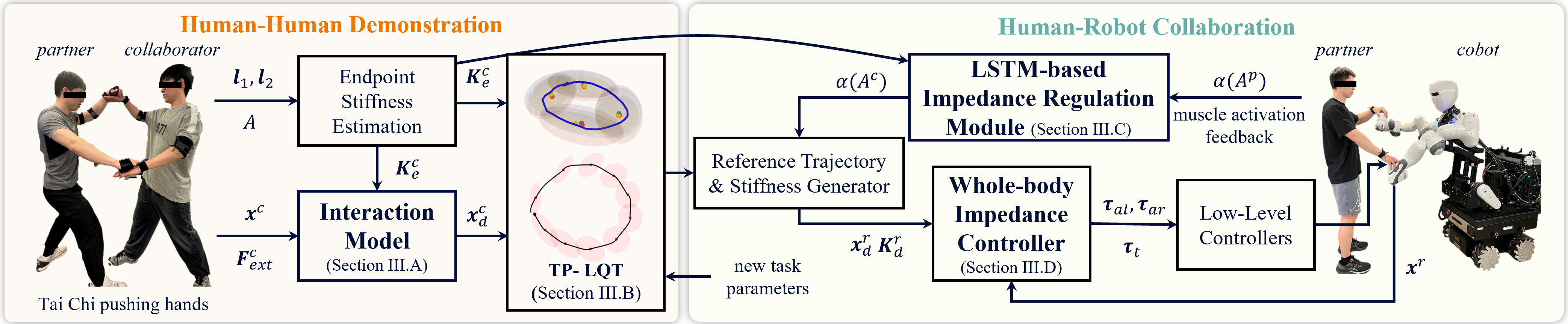} 
\caption{Multi-modal data acquisition records synchronized human motion and interaction signals, taking as inputs the joint/endpoint poses, EMG-based muscle activations, and interaction forces. Endpoint stiffness estimation uses human kinematics and muscle activation levels as inputs to estimate the human endpoint stiffness, providing impedance profiles for learning and regulation. The interaction model takes the measured TCP trajectory, the estimated stiffness, and the interaction force as inputs to compute the human reference trajectory consistent with a spring–damper relationship. The TP-LQT module consumes the human reference trajectory, the human stiffness profile, and task parameters to generate the robot reference trajectory and the robot stiffness profile. The LSTM-based impedance regulation module takes the collaborator’s muscle co-contraction as input and outputs a robot impedance scaling index for online adaptation. Finally, the whole-body impedance controller receives the robot reference trajectory, the current robot TCP trajectory, and the reference impedance parameters as inputs and outputs joint torques for the arm and torso to execute the task on CURI.}
\label{Fig.framework}\vspace{-3mm}
\end{figure*}

Two dominant upper arm muscles biceps brachii (BB) and triceps brachii (TB) are chosen to calculate the co-contraction index $\alpha(A)$, which is illustrated as:
\begin{equation}\label{cocontraction}
  A = \frac{A_{BB} + A_{TB}}{2}, 
\end{equation}
where $A_{BB}$ and $A_{TB}$ represent the muscle activation levels of BB and TB, respectively. Assume that the EMG signals from the demonstrator are collected as $[U_{BB}, U_{TB}] \in \mathbb{R}^{2}$ with $2$ EMG sensors in measurement at each time step. They are further processed by filters and normalized using Maximal Voluntary Contraction (MVC). Then, the muscular activation levels based on the MVC and the processed EMG data can be calculated as $[A_{BB}, A_{TB}] \in \mathbb{R}^{2}$. 

The parameters $a_1$, $a_2$, $b_1$, and $b_2$, which vary among individuals, are identified using an optimization method that minimizes the error between the estimated and actual endpoint stiffness matrices. A perturbation method is employed to measure the matrices \cite{arashijrr}. Upon identifying these parameters for each demonstrator, we can estimate their endpoint stiffness matrices during demonstrations and learn representations of impedance regulation skills.

\subsection{Interaction Force-Based Stiffness Estimation}\label{interaction force-based stiffness}
In addition to EMG-based stiffness estimation, the end-effector impedance can be observed online based on the interaction force and position. Since offline identification is sensitive to stochastic disturbances, the stiffness-damping model can be regarded as a latent state and updated via an Extended Kalman Filter (EKF), which linearises the interaction dynamics and fuses force–position residuals for real-time parameter adaptation. By referring to the observer design in \cite{zhang2023observer}, a state vector comprising Cartesian stiffness and damping, along with a measurement model, can achieve the stiffness estimation.

\section{Methodology} \label{learning}
In this section, the overall framework (as shown in Fig. \ref{Fig.framework}) will be presented. We first introduce an interaction model that represents the contact dynamics between the robot end-effector and the environment, where `environment’ denotes the external physical counterparts in contact during the specific task (e.g., manipulated objects and/or the human collaborator). The endpoint stiffness matrices along the endpoint trajectories are estimated and the reference trajectories are then calculated with data recorded from human demonstrators. A TP-GMM is proposed for learning motion and impedance representation from the demonstrations by corresponding the reference trajectory and the stiffness profiles. The LQT method for trajectory planning and impedance variation is then raised afterwards for robot execution. Meanwhile, an impedance regulation module to learn the compliance relationship between two demonstrators is proposed. Finally, a whole-body impedance controller is applied to perform the reference trajectory and impedance variables for the CURI. 

\subsection{Interaction Model}\label{interaction model}
The interaction model between the cobot and the environment encapsulates the dynamics of the physical behaviors during the collaboration, which is supposed to imitate human collaboration skills by utilizing both the Tool Center Point (TCP) trajectory $\bm{x}^c$ and force information $\bm{F}^c_{ext}$. Hence, we propose to learn the reference trajectory $\bm{x}^c_{d}$ and the stiffness matrix of the interaction model between the human and the environment during the human-human demonstrations and then transfer to the cobot for HRC.

Considering the movement of the robot/human TCP ($\bm{x}^r$/$\bm{x}^c$, represented as $\bm{x}$ in the following formulation) as a point-mass moving in Cartesian space under the effect of a control input and interaction force. To achieve the desired dynamics of the tasks, the problem can be formulated as designing the appropriate control input. Specifically, the Cartesian-space impedance model to dynamically connect the positions, velocities and accelerations with the interactive forces is:
\begin{equation}\label{impedance model}
  \bm{\Lambda}_{d}\bm{\ddot{\widetilde{x}}} + \bm{D}_{d}\bm{\dot{\widetilde{x}}} + \bm{K}_{d}\bm{\widetilde{x}}
    = \bm{F}_{ext}, 
\end{equation}
where $\bm{\widetilde{x}} = \bm{x}_{d} - \bm{x} \in \mathbb{R}^{6}$. $\bm{\Lambda}_{d}, \bm{D}_{d}, \bm{K}_{d} \in \mathbb{R}^{6 \times 6}$ are the symmetric and positive definite matrices of the desired inertia, damping, and stiffness, respectively. $\bm{F}_{ext}$ is the external force applied to the TCP.

As mentioned above, we propose to learn the reference trajectory for the cobot to execute. However, we can only observe the human TCP trajectory $\bm{x}^c$ of the demonstrations instead of the variable $\bm{x}^c_d$. Assuming the velocity and acceleration of the reference trajectory remain at 0 \cite{rozo2016learning}, the dynamic model can be rewritten based on (\ref{impedance model}):
\begin{equation}\label{eq3}
  \bm{\Ddot{x}} = \bm{K}_d (\bm{x}_d - \bm{x}) - \bm{D}_d \bm\dot{{x}}  + \bm{F}_{ext}.
\end{equation}

The reference trajectory $\bm{x}_d$ can be calculated accordingly if TCP pose $\bm{x}$, stiffness (damping) matrix $\bm{K}_d$ ($\bm{D}_d$), and external forces $\bm{F}_{ext}$ are known. 

\subsection{Motion Representation and Planning}\label{trajectory planning}
Human reference trajectory and impedance profiles are modeled using TP-GMM, derived from human demonstrations and transferred to the cobot by modifying task parameters. A smooth reference trajectory, synchronized with the impedance variables, is then generated using an LQT-based method for robotic execution. This integrated approach of motion representation and planning is termed TP-LQT (Fig. \ref{Fig.framework}). 

The reference trajectory of each demonstration is firstly calculated by referring to (\ref{eq3}). The newly generated trajectories can be observed from different frames and GMM is adopted to cluster points on the demonstration into multiple Gaussian distributions in each frame $i \in [1, 2]$. Compared with previous works that applied TP-GMM \cite{calinon2016tutorial}, our method binds motion and impedance profiles and then fits a joint GMM over a concatenated feature space (7D pose $+$ 3D stiffness). After that, a motion with a corresponding impedance profile considered the task parameter can be reproduced via the product of experts (PoE). The task parameter denotes the task-specific frames that condition the TP-GMM. For box transportation, the start/goal poses (and induced height/distance) capture new start–goal configurations. For Tai Chi pushing hands, partner-aligned frames derived from each collaborator’s kinematics capture new collaborator reach and workspace. Note that the task parameter is defined as the transformation matrices of the start point and endpoint of multiple demonstrations $\boldsymbol{\zeta}$ from the world origin, namely $\{\boldsymbol{b}^{(i)}, \boldsymbol{A}^{(i)}\}_{i=1, 2}$. We first use these matrices to transform and align demonstrations with respect to each observer frame $\boldsymbol{X}_t^{(i)}={\boldsymbol{A}^{(i)}}^{-1}(\boldsymbol{\zeta}_t-\boldsymbol{b}^{(i)})$. Then we use a TP-GMM with $J$ Gaussian distributions $\{\boldsymbol{\mu}_j^{(i)}, \boldsymbol{\Sigma}_j^{(i)}\}_{i=1, 2}, j\in[1, J]$ to represent $\boldsymbol{X}_t^{(i)}$ in each frame. Thus, the reproduction is expected to lie within the distributions $\mathcal{N}\left(\hat{\boldsymbol{\xi}}^{(i)}_{j}, \hat{\boldsymbol{\Sigma}}^{(i)}_{j}\right)_{i=1, 2}$, where $\hat{\boldsymbol{\xi}}^{(i)}_j=\boldsymbol{A}^{(i)} \boldsymbol{\mu}^{(i)}_j+\boldsymbol{b}^{(i)}$, $\hat{\boldsymbol{\Sigma}}^{(i)}_j=\boldsymbol{A}^{(i)} \boldsymbol{\Sigma}^{(i)}_j {\boldsymbol{A}^{(i)}}^{\top}$.

The transformation matrices of the start point and endpoint are changed to generate the motion in new situations, hence a new GMM is generated by PoE:
\begin{equation}
\mathcal{N}\left(\hat{\boldsymbol{\xi}}_{j}, \hat{\boldsymbol{\Sigma}}_{j}\right) \propto \prod_{i=1}^2 \mathcal{N}\left(\hat{\boldsymbol{\xi}}_{j}^{(i)}, \hat{\boldsymbol{\Sigma}}_{j}^{(i)}\right),
\end{equation}
with the result as  $\hat{\boldsymbol{\Sigma}}_{j}=\left(\sum_{i=1}^P \hat{\boldsymbol{\Sigma}}_{j}^{(i)^{-1}}\right)^{-1}$, $\hat{\boldsymbol{\xi}}_{j}=\hat{\boldsymbol{\Sigma}}_{j} \sum_{i=1}^P \hat{\boldsymbol{\Sigma}}_{j}^{(i)^{-1}} \hat{\boldsymbol{\xi}}_{j}^{(i)}$. Thus, the reproduced motion $\hat{\boldsymbol\zeta}$ from this GMM should follow the style of demonstrations.

TP-GMM focuses on generating a reference trajectory $\hat{\boldsymbol\zeta}$ with desired task parameters by assuming the cobot has an ideal trajectory controller. Thus, a controller that considers the impedance and reference trajectory is needed to generate the motion in new situations. The tracking cost function of LQT method is defined as:
\begin{equation}
C=\sum_{j=1}^2\left(\hat{\boldsymbol{\xi}}_{k, t}^{(j)} -\hat{\boldsymbol{\zeta}}\right)^{\top} \hat{\boldsymbol{\Sigma}}_{k, t}^{(j)-1} \left(\hat{\boldsymbol{\xi}}_{k, t}^{(j)} - \hat{\boldsymbol{\zeta}}\right) + \boldsymbol{U}^{\top} \boldsymbol{R} \boldsymbol{U},
\end{equation}
where $\boldsymbol{R}$ is the control cost matrix and $\boldsymbol{U}$ is the control input. A two-step optimization is used to solve the problem to obtain the trajectory $\hat{\boldsymbol{\xi}}$ and the input:
\begin{equation}
\setlength{\abovedisplayskip}{3pt}
\setlength{\belowdisplayskip}{3pt}
\begin{aligned}
\hat{\boldsymbol{\xi}} & =\arg \min_{\boldsymbol{\xi}} \sum_{j=1}^2\left(\hat{\boldsymbol{\xi}}_{k, t}^{(j)}-\hat{\boldsymbol\zeta}\right)^{\top} {\hat{\boldsymbol{\Sigma}}_{k, t}^{(j)^{-1}}} \left(\hat{\boldsymbol{\xi}}_{k, t}^{(j)}-\hat{\boldsymbol\zeta}\right), \\
\hat{\boldsymbol{U}} & =\arg \min_{\boldsymbol{U}}\left(\hat{\boldsymbol{\xi}}-\boldsymbol{\zeta}\right)^{\top} {\hat{\boldsymbol{\Sigma}}}^{-1}\left(\hat{\boldsymbol{\xi}}-\boldsymbol{\zeta}\right)+\boldsymbol{U}^{\top} \boldsymbol{R} \boldsymbol{U}.
\end{aligned}
\end{equation}

\subsection{Impedance Regulation Skill Learning} \label{impedance estimation method}
In this part, a regulatory factor will be defined to describe the compliance reaction based on the partner, which is learned from human-human demonstrations. By substituting (\ref{eigendecomposition}) into (\ref{human impedance relationship}), we have:
\begin{equation}\label{relationship}
\alpha(A^c) \bm{V}^c \bm{H}_s^c {\bm{V}^c}^T = h( \alpha(A^p) \bm{V}^p \bm{H}_s^p {\bm{V}^p}^T),
\end{equation}
where $\bm{V}^c$, $\bm{V}^p$, $\bm{H}_s^c$, and $\bm{H}_s^p$ represent the orthonormal matrices and the diagonal matrices containing the configuration dependence eigenvalues of the collaborator and the partner, respectively. The arm configurations of two individuals remain corresponding during human demonstrations, which makes the variation of $\bm{V}$ and $\bm{H}_s$ basically unchanged along the movement from one demonstration to another. Hence, the variation of the regulatory factor during tasks will be learned between the two synergistic contributions of the muscle co-contractions $\alpha(A^c)$ and $\alpha(A^p)$.

Machine learning has a high potential to build the implicit model between two individuals in collaboration, especially for those tasks with periodicity \cite{mukherjee2022survey}. We propose to use a Long-Short-Term Memory (LSTM) neural network for regulation policy learning since the data of muscular synergistic contributions are strictly based on time series during the demonstrations \cite{semeraro2023human}. These kinds of time series prediction models have been applied to model the interaction behaviors, such as EMG-to-force \cite{xie2024deep} and human intention recognition \cite{9568762, schirmer2023}. To achieve this, the data will constitute the form with the LSTM function as $h: X \rightarrow Y$. A loss function will be defined to evaluate the performance of the model by comparing the predicted outputs with the actual results. 

In this paper, the LSTM network model is trained using input data consisting of the partner's muscle co-contraction signals and the corresponding demonstrated trajectories, while the output is designed to predict the collaborator's muscle co-contraction. Hence, the form with the LSTM function as $h: (\alpha(A^p), \boldsymbol{\zeta}) \rightarrow \alpha(A^c)$. This design enables the model to capture the temporal relationships between partner behaviors and task dynamics. By learning this mapping, the trained model can generalize to different trajectories, which may arise due to variations in individual personalities or movement styles, thereby enhancing its adaptability to diverse interaction scenarios. LSTM is sensitive to the scale of the input data, especially when using the sigmoid activation function. Therefore, we normalize the raw data to a range of $[0, 1]$ as inputs and de-normalize the predictive outputs from the model for better training. We connect multiple sets of demonstration data end-to-end to form the dataset for model construction. 

The effectiveness of this design is validated in Section IV, where the proposed LSTM-based model demonstrates its capability to adapt to various interaction scenarios. Specifically, the experiments show that the model successfully generalizes to diverse trajectories, accommodating differences in individual movement styles.

\subsection{Whole-Body Impedance Controller} \label{whole-body} 
A whole-body Cartesian impedance controller of CURI is introduced to achieve human-robot collaboration. CURI is a Dual-Arm mobile platform (as shown in Fig. \ref{Fig.framework}) with one velocity-controlled 3-DoFs Robotnik SUMMIT-XL STEEL mobile base, one velocity-controlled 3-DoFs self-designed torso, and two torque-controlled 7-DoFs Franka Emika Panda robotic arms. In this paper, the robotic arms and the torso were considered and the corresponding whole-body impedance controller was designed.

A force-torque interface is preferred since we target to implement a whole-body impedance control law. Hence, we design Cartesian admittance controllers based on the velocity interface for the torso. Considering that the torso applies high-gain velocity controllers to realize the admittance dynamics. All disturbances including the external dynamic effect from the arms and the coupling dynamics from the torso are assumed to be compensated by these controllers. 

The whole-body decoupled dynamics can be written as:
\begin{equation}\label{whole-body-dynamic}
\begin{small}
\setlength{\arraycolsep}{0.5pt}
\begin{aligned}
    &\left ( \begin{matrix}
    \bm{M}_{t}^{adm} & \bm{0} & \bm{0}\\
      \bm{0} & \bm{M}_{al} & \bm{0}\\
      \bm{0} & \bm{0} & \bm{M}_{ar}
    \end{matrix} \right ) \left ( \begin{matrix}
      \bm{\Ddot{q}}_{t}\\
      \bm{\Ddot{q}}_{al} \\
      \bm{\Ddot{q}}_{ar} 
    \end{matrix} \right ) + \left ( \begin{matrix}
      \bm{0}\\
      \bm{G}_{al}\\
      \bm{G}_{ar}
    \end{matrix}\right ) +\\
    &\left ( \begin{matrix}
        \bm{D}_{t}^{adm} & \bm{0} & \bm{0}\\
        \bm{0} & \bm{C}_{al} & \bm{0}\\
        \bm{0} & \bm{0} & \bm{C}_{ar}
    \end{matrix} \right )\left ( \begin{matrix}
      \bm{\dot{q}}_{t}\\
      \bm{\dot{q}}_{al}\\
      \bm{\dot{q}}_{ar}
    \end{matrix}\right )
    = \left ( \begin{matrix}
      \bm{\tau}_{t}^{vir}\\
      \bm{\tau}_{al}\\
      \bm{\tau}_{ar}
    \end{matrix}\right ) + \left ( \begin{matrix}
      \bm{\tau}_{t}^{ext}\\
      \bm{\tau}_{al}^{ext}\\
      \bm{\tau}_{ar}^{ext}
    \end{matrix}\right ),
\end{aligned}
\end{small} 
\end{equation}
where $\bm{\dot{q}}_{t} \in \mathbb{R}^{3}$ represent the velocities expressed in the joint space of the torso. $\bm{M}_{t}^{adm} \in \mathbb{R}^{3\times 3}$ and $\bm{D}_{t}^{adm} \in \mathbb{R}^{3\times 3}$ are respectively the virtual inertial and virtual damping set for the torso in the admittance controllers. $\bm{\tau}_{t}^{vir} \in \mathbb{R}^{3}$ and $\bm{\tau}_{t}^{ext} \in \mathbb{R}^{3}$ are respectively the commanded and the external torques for the torso. $\bm{q}_{al}, \bm{q}_{ar} \in \mathbb{R}^{7}$ represent the joint angles of the left robotic arm and the right robotic arm, respectively. $\bm{M_{al}}, \bm{M_{ar}} \in \mathbb{R}^{7 \times 7}$, $\bm{C_{al}}, \bm{C_{ar}} \in \mathbb{R}^{7}$, and $\bm{G_{al}}, \bm{G_{ar}} \in \mathbb{R}^{7}$ are the inertia matrix, gravity term, Coriolis and centrifugal terms for the two arms, respectively. $\bm{\tau}_{al}, \bm{\tau}_{ar} \in \mathbb{R}^{7}$ and $\bm{\tau}_{al}^{ext}, \bm{\tau}_{ar}^{ext} \in \mathbb{R}^{7}$ are the commanded and the external torques for the robotic arms, respectively. 

CURI has two Tool Center Points (TCPs) corresponding to the left and right robotic arm, respectively. Hence, the CURI can be split into two chains based on the two TCPs with shared parts (torso). Equation (\ref{whole-body-dynamic}) can be rewritten as:
\begin{equation}\label{decoupled dynamics}
  \bm{M}_{i}(\bm{q}_{i})\bm{\ddot{q}}_{i} + \bm{C}_{i}(\bm{q}_{i}, \bm{\dot{q}}_{i})\bm{\dot{q}}_{i} + \bm{G}_{i}(\bm{q}_{i})
    = \bm{\tau}_{i} + \bm{\tau}_{i}^{ext}, \quad  i = l, r
\end{equation}
where $\bm{q}_{i} = \left ( \begin{matrix} \bm{q}_{t}^T \ \bm{q}_{ai}^T \end{matrix} \right )^T$ represents the joint angles vector of the chain $i$, $\bm{M}_{i}(\bm{q}_{i})$ is the corresponding joint-space inertia matrix, $\bm{C}_{i}(\bm{q}_{i}, \bm{\dot{q}}_{i})$ is the Coriolis/centrifugal matrix, and $\bm{G}_{i}(\bm{q}_{i})$ the vector of gravity. $\bm{\tau}_{i}$ and $\bm{\tau}_{i}^{ext}$
represent the corresponding control input and external torque in joint space, respectively.

Once the robot dynamic (\ref{decoupled dynamics}) and impedance model (\ref{impedance model}) are given, the Cartesian impedance controller proposed in \cite{ott2008cartesian} can be extended to CURI, while its stability can also be guaranteed. The whole-body impedance controller generating high-level torque command for each chain $\tau_{i}$ ($i = l, r$) can be expressed as:
\begin{equation}\label{whole-body-impedance-controller}
\begin{aligned}
    \left ( \begin{matrix}
      \bm{\tau}_{t}^T &
      \bm{\tau}_{ai}^T
    \end{matrix} \right )^T = \bm{\tau}_{i}^{task} + \bm{\tau}_{i}^{null}.
\end{aligned}
\end{equation}

The $\bm{\tau}_{i}^{task}$ corresponding to the given main task is:
\begin{equation}\label{whole-body-impedance-controller-main-task}
    \bm{\tau}_{i}^{task} = \bm{J}_{i}^{T}(-\bm{K}_{d,i}(\bm{{x}_{i}}-\bm{{x}_{d,i}})-\bm{D}_{d,i}\bm{\dot{x}}_{i}),
\end{equation}
where $\bm{J}_{i} \in \mathbb{R}^{6 \times 10}$ represents the Jacobian matrix corresponding to each chain. 

\begin{figure*}[t]
\centering
\includegraphics[width=0.98\textwidth]{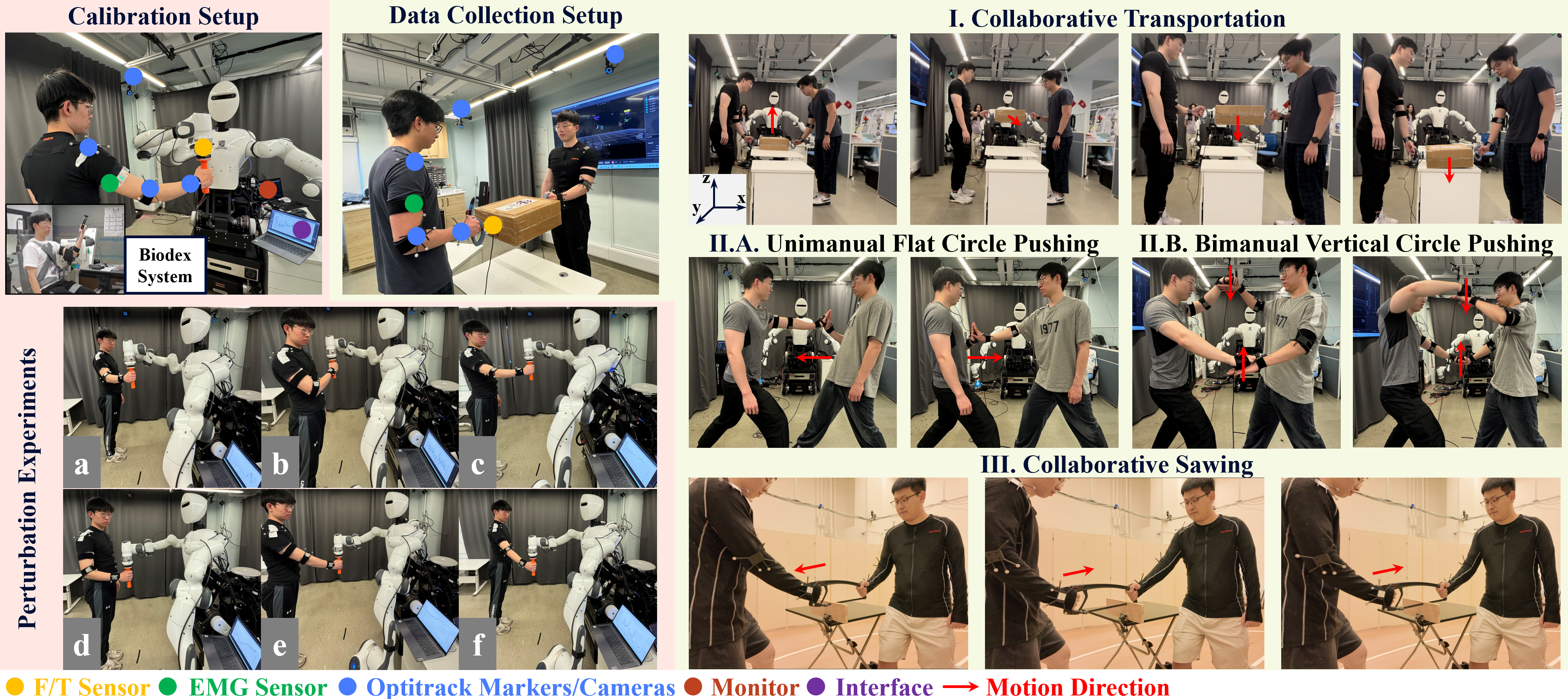} \vspace{-4mm}
\caption{\textit{Calibration setup:} endpoint stiffness parameters calibration through perturbation methods. \textit{Data collection setup:} two subjects were attached to the OptiTrack markers on their arm joints while the co-contraction muscle signals were recorded by EMG sensors. F/T sensors were attached to the object/subject. \textit{Perturbation experiments:} six different human arm configurations $a-f$ were selected for calibration through physical human-robot interaction. The snapshots of human-human demonstrations. \textit{I:} collaborative transportation, \textit{II.A:} Tai Chi unimanual flat circle pushing, \textit{II.B:} Tai Chi bimanual vertical circle pushing, and \textit{III:} collaborative sawing.} 
\label{Fig.demonstration_transportation} \vspace{-2mm}
\end{figure*}

The null-space task input is defined as:
\begin{equation}\label{null_space_torque}
    \bm{\tau}_{i}^{null} = (\bm{I} - \bm{J}_{i}^T\bm{J}_{i}^{+T})\bm{\tau}_{i}^{0},
\end{equation}
where $\bm{\tau}_{i}^{0}$ represents the corresponding torque projected in the null space of the main task, $\bm{N}(\bm{q}_{i}) = \bm{I} - \bm{J}_{i}^T\bm{J}_{i}^{+T} \in \mathbb{R}^{10 \times 10}$ is the projection matrix in order to prevent interference with the Cartesian impedance behavior of the main task. $\bm{J}_{i}^{+}$ expresses the generalized inverse of $\bm{J}_{i}$.

Finally, after obtaining the designed torques $\bm{\tau}_{i} = \left ( \begin{matrix} \bm{\tau}_{t}^T \ \bm{\tau}_{ai}^T \end{matrix} \right )^T$, $\bm{\tau}_{ai}$ is sent to the robotic arm directly through its torque command interface. As the torso received velocity command only, the $\bm{\tau}_{t}$ can generate the velocity command $\bm{\dot{q}}_{t}$ through the corresponding admittance interfaces in (\ref{whole-body-dynamic}).

The above formulation allows us to control one chain of CURI at a time. However, it may cause conflicts when we are supposed to control both chains at the same time since the torso are shared parts. To address this problem, we propose a leader-follower method regarding the two chains. Assuming the left arm as the \textit{leader arm}, the whole-body impedance controller for its corresponding chain can be designed with $\bm{\tau}_{l}^{0}$ in (\ref{null_space_torque}), which is generated by some predefined sub-tasks, i.e., optimizing the joint configuration for the robotic arm \cite{dietrich2016whole}. This paper defines the sub-task as the torso movement strategy based on human demonstrations. Specifically, the cobot torso is supposed to move in the same direction as the demonstrator's trunk at different stages of the task \cite{neo2007whole}. Then the calculated $\bm{\tau}_{t}$ and $\bm{\tau}_{al}$ of the \textit{leader arm} can be treated as a sub-task of the controller corresponding to the right chain of CURI, whose corresponding arm is referred to as the \textit{follower arm}. The $\bm{\tau}_{r}^{0}$ of the \textit{follower arm} can be chosen as:
\begin{equation}
    \bm{\tau}_{r}^{0} = \left ( \begin{matrix}
  \bm{\tau}_{t}^T, \
  \bm{0}_{1\times 7}
\end{matrix} \right ).
\end{equation}

This impedance model enables the cobot to perform human-like behavior adjustment by varying damping and stiffness while tracking the reference trajectory based on task requirements. To ensure stability, passivity conditions for time-varying impedance parameters must be satisfied \cite{kronander2016stability}. When these conditions are met, the calculated impedance variables $\bm{K}_d^r$ and $\bm{D}_d^r$ can be fed into the controller to achieve the desired behaviors.

\section{Experiments}\label{V}

Three scenarios including four typical tasks were introduced to verify the performance of our framework under the leader-follower mode and the mutual adaptation mode, which are shown in Fig. \ref{Fig.demonstration_transportation}. Five human subjects were involved with four males and one female (age: $28 \pm 2$ years; height: $1.74 \pm 0.14$ $m$). Comparative experiments of our proposed HI-ImpRSL, a fixed impedance control method (FIC) \cite{9362246}, an EMG-based variable impedance control method (EMG-based VIC) \cite{peternel2016adaptation}, a neural network-based variable impedance skill learning framework (NN-based VIC) with interaction force-based stiffness estimation \cite{zhang2023neural}, and a TP-LQT based variable impedance control method (HI-ImpRSL w/o LSTM) \cite{rozo2016learning}, illustrate the advantages of our HI-ImpRSL method.

\subsection{Endpoint Stiffness Parameter Calibration}
Perturbation experiments are designed to identify the parameters of each demonstrator, which are applied to extract the endpoint stiffness during human-human demonstrations. The calibration setup is illustrated in Fig. \ref{Fig.demonstration_transportation}. We used CURI arm for parameter identification and mounted a force/torque (F/T) sensor (SRI, 100 $Hz$) connecting the handle and the end effector. An optical system (OptiTrack, 120 $Hz$) was applied to track the poses of the shoulder, elbow, and wrist by attaching markers to these human joints. Besides, a wireless EMG device (Delsys Trigno, 2000 $Hz$) and a monitor were used to visualize the real-time muscle co-contraction. 

We selected biceps brachii (BB) and triceps brachii (TB) as antagonistic pairs to compute co-contraction (Sec. \ref{emg-based stiffness}). EMG was acquired with a Delsys Trigno system at 2000 $Hz$, followed by a fourth-order 100 $Hz$ high-pass Butterworth filter to clean the raw data. Signals were full-wave rectified and processed for the envelope at 250 $Hz$, each muscle’s activation was normalized to its MVC obtained with a Biodex system, and activations were scaled to $[0,1]$. The resulting activations were used to compute co-contraction in (\ref{cocontraction}).

During perturbation experiments, a trajectory with perturbations in six directions ($x$, $y$, $z$, $xy$, $xz$, $yz$) was given for each test. The order in which these perturbations appear was randomized to avoid subject adaptation. The peak displacement of each perturbation in one trial was set as 0.02 $m$ while the duration was 0.5 $s$. We have tested six arm configurations with three different muscle co-contraction levels of each configuration, which are minimum co-contraction, 20\% MVC, and 40\% MVC. Hence, the total number of trials for each subject was 18 during stiffness parameters calibration. The parameters ${a_1, a_2, b_1, b_2}$ of five subjects in this paper were given as follows in Table \ref{tab1}. 

\begin{table}[h]
\caption{Human Endpoint Stiffness Parameters}
\begin{center}
\begin{tabular}{c|cccc}
\hline
\small{Subject} & \small{\textit{$a_1$}} &  \small{\textit{$a_2$}} &  \small{\textit{$b_1$}} &  \small{\textit{$b_2$}} \\
\hline
\small{S1} & \small{0.272} &  \small{1.314} & \small{3847.141} & \small{151.684}   \\
\small{S2} & \small{0.107} & \small{2.200} & \small{2678.765} & \small{149.597}  \\
\small{S3} & \small{0.399} & \small{2.926} & \small{1819.695} & \small{128.581}  \\
\small{S4} & \small{0.341} & \small{4.073} & \small{2699.123} & \small{112.562}  \\
\small{S5} & \small{0.167} & \small{4.528} & \small{1260.290} & \small{94.951}  \\
\hline
\end{tabular}
\label{tab1}
\vspace{-4mm}
\end{center}
\end{table}

\begin{figure*}[t]
\centering
\setlength{\abovecaptionskip}{0.cm}
\includegraphics[width=0.98\textwidth]{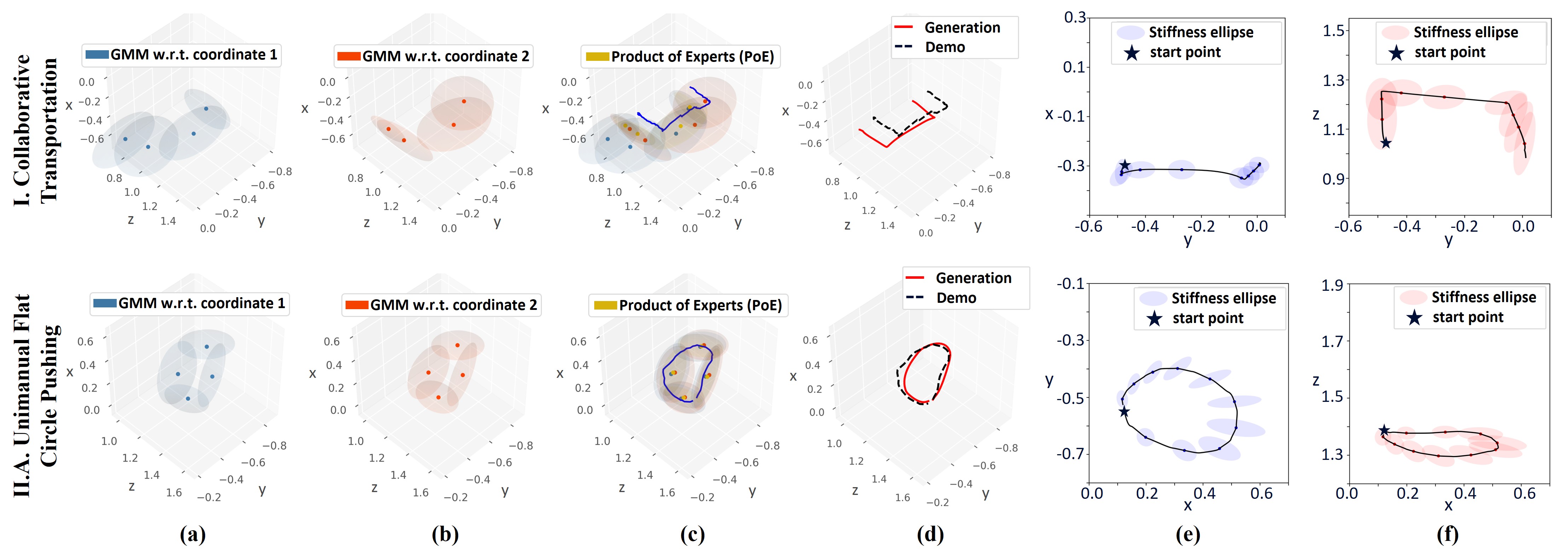} 
\caption{Learning results for motion/impedance representation and generation by TP-LQT method are illustrated. GMM is firstly adopted in coordinates 1 and 2 to cluster points into multiple Gaussian distributions and represent the demonstrations with PoE \textit{(a-c)}. By considering the task parameter with a new situation, a trajectory is generated via PoE \textit{(d)}. Finally, the corresponding stiffness ellipses are drawn along the generated trajectory in planes with the LQT method \textit{(e-f)}. \textit{I:} collaborative transportation, \textit{II.A:} Tai Chi unimanual flat circle pushing. } 
\label{Fig.tpgmm_boxcarrying}\vspace{-5mm}
\end{figure*}

\begin{figure*}[!tbp]
\centering
\includegraphics[width=0.95\linewidth]{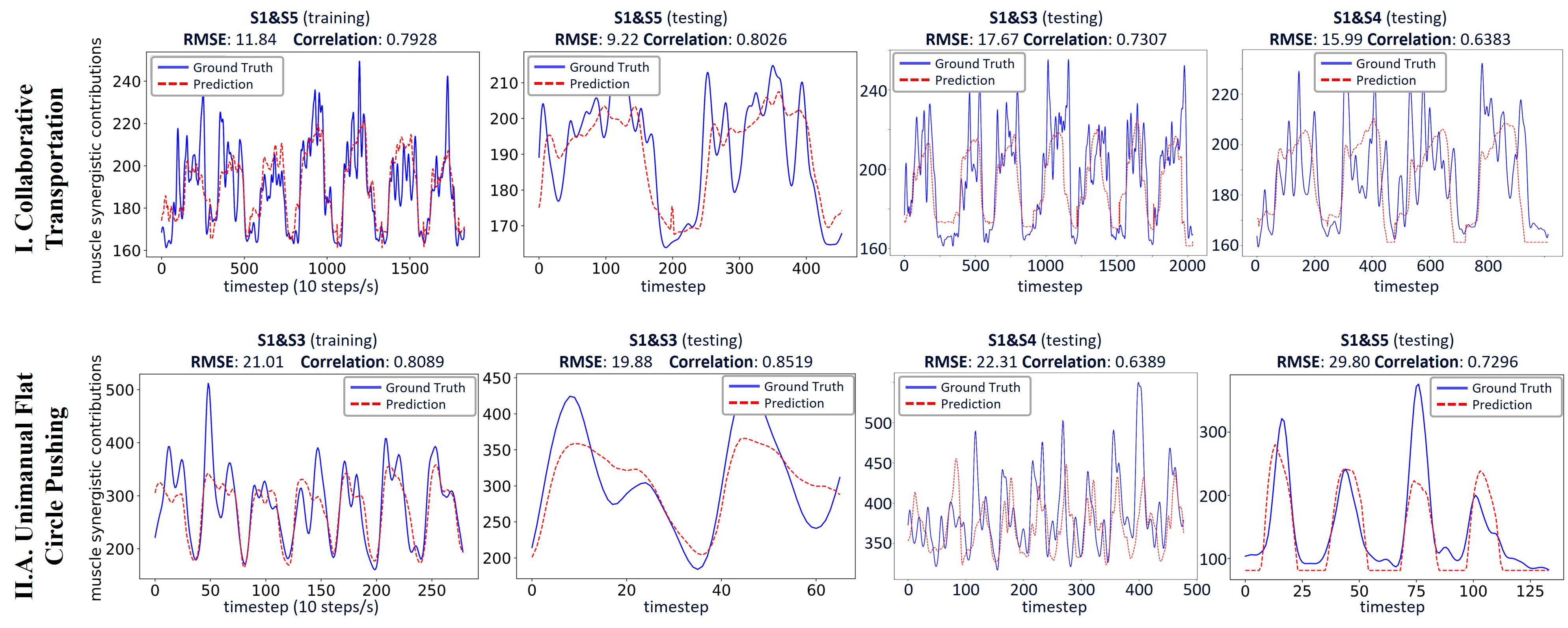} 
\caption{LSTM model training and testing for regulatory factor learning. \textit{I:} collaborative transportation, \textit{II.A:} Tai Chi unimanual flat circle pushing.}
\label{Fig.lstm_box} \vspace{-4mm}
\end{figure*}

\subsection{Collaborative Transportation}
\subsubsection{Task Definition}

The first scenario is human-robot collaborative transportation, where the cobots are expected to help the human partners by manipulating and transporting the objects while dealing with the position and impedance constraints (as shown in Fig. \ref{Fig.demonstration_transportation}.I). Specifically, human and robot will collaboratively lift, translate, and lower the object to the target pose under leader-follower mode. It is worth noting that the start and the end poses of the object are not fixed during collaborations. In this task, we propose to teach the cobot follower to perform consistent compliance behavior as the human leader (shown in Fig. \ref{Fig.physical collaboration skill}).

One arm and the torso of CURI were used for collaborative transportation tasks. A robotic hand from qb SoftHand2 Research was applied as the end effector for grasping. A whole-body Cartesian impedance control interface was implemented based on section \ref{whole-body} in ROS. By giving the commanded impedance matrices and the reference trajectory with an update frequency of 100 $Hz$, the controller outputted the commanded torques of each joint for the robotic arms and calculated joint velocities for torso joints to generate expected motions with a frequency of 1000 $Hz$. Note that the gravity compensation was redesigned due to the unique installations of the arms. 

\begin{figure*}[!t]
\centering
\includegraphics[width=0.95\textwidth]{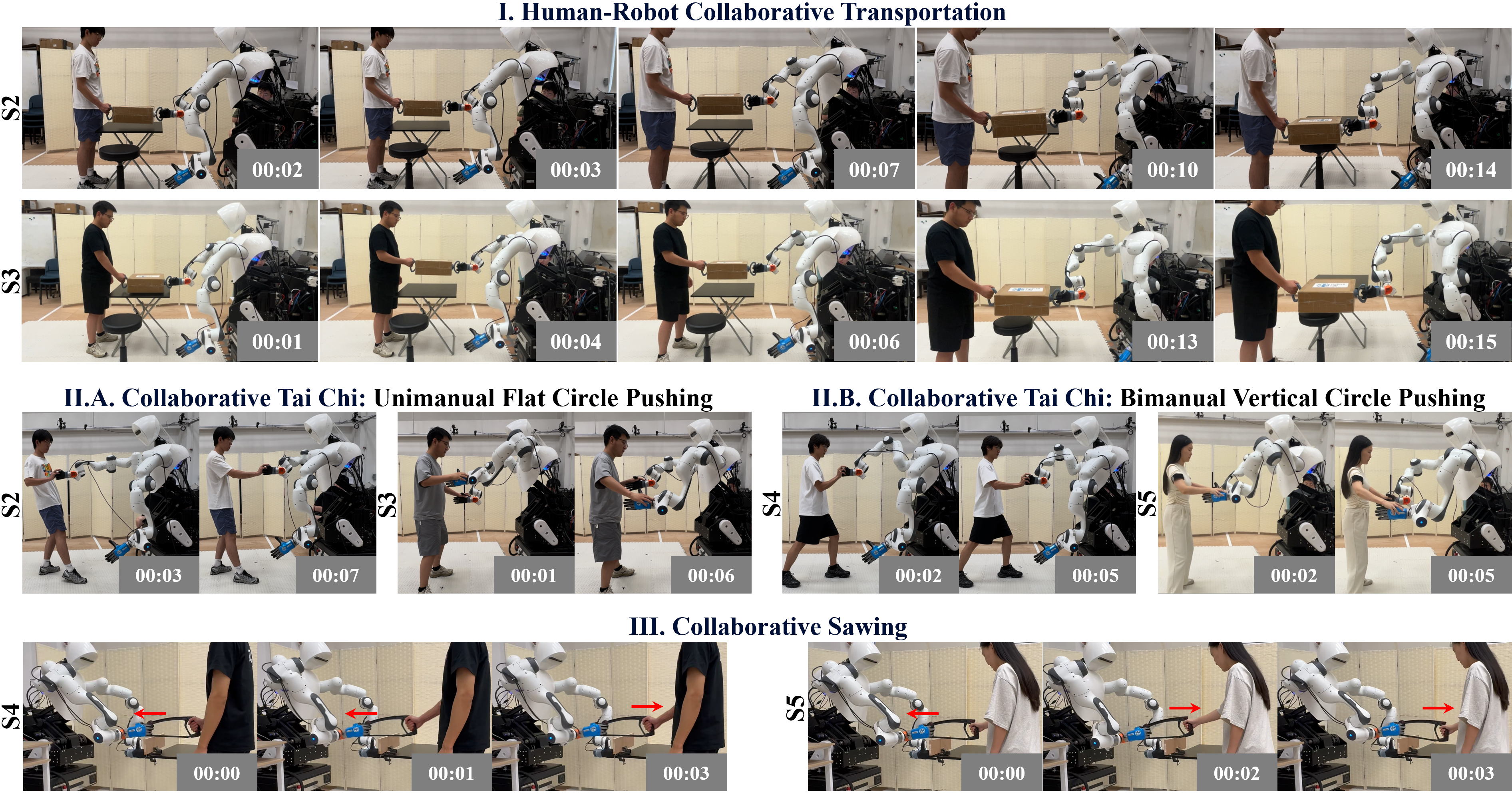} \vspace{-2mm}
\caption{The snapshots of human-robot collaborative tasks by HI-ImpRSL method. \textit{I:} object transportation, \textit{II:} Tai Chi pushing hands, and \textit{III:} sawing.}
\label{Fig.exp_overall} \vspace{-2mm}
\end{figure*}

\subsubsection{Trajectory and Impedance Generation from Demonstration}
The demonstration data of the collaborative transportation were recorded by the OptiTrack system, EMG sensors, and force sensors with the corresponding software. The demonstrators' motions were obtained from the OptiTrack markers, which were fixed on the human joints (shown in Fig. \ref{Fig.demonstration_transportation}). The data were used for offline learning to generate a proper representation of the cobot motions. Besides, the demonstrators were wearing two EMG sensors on the upper body (same positions as the calibration process) to record the muscle activities. The recording and processing procedures of EMG signals were the same as the calibration part. These data were used for human endpoint stiffness estimation. The reference trajectories were then calculated based on the human endpoint motion, stiffness, and interactive force by (\ref{eq3}). 

As shown in Fig. \ref{Fig.tpgmm_boxcarrying}.I, we utilized three calculated reference trajectories of human demonstrations in GMM. The result of the representation with regard to the PoE was illustrated for the object transportation task. Furthermore, by adding the task parameter and extending to TP-GMM, the motion reproduction was achieved with new initial and final poses of the reference trajectory and the generated reference trajectory was given. Besides, the corresponding stiffness variables along the trajectory were calculated based on the LQT method. 

The LSTM model was presented in the upper part of Fig. \ref{Fig.lstm_box}, which compares the predicted muscle synergistic contributions with the ground truth values of the collaborator during an object transportation task. The demonstration data from Subject 1 (S1, collaborator) and Subject 5 (S5, partner) were used as the training set for constructing the model, while additional subject combinations (S1 and S3, S1 and S4) were utilized as testing sets. The accuracy of the LSTM network predictions was evaluated using the Root Mean Squared Error (RMSE) and the correlation coefficient. The results indicate that the prediction trends align well with the ground truth, with performance metrics for the testing sets (S1\&S3: RMSE 17.67, correlation 0.7307; S1\&S4: RMSE 15.99, correlation 0.6383) slightly lower than those of the training set (RMSE 11.84, correlation 0.7928). The observed performance decline in the testing sets can be attributed to the inherent variability of EMG signals across different individuals. Given these differences, such a reduction in prediction accuracy is acceptable as long as the model successfully captures the overall trends, which is crucial for generalizing to diverse interaction scenarios.

\subsubsection{Experimental Results}
In real world human-robot experiments shown in Fig. \ref{Fig.exp_overall}, the human-like cobot, CURI, was equipped with an OptiTrack marker behind its torso to obtain the poses of the end effector in the world coordinates. These data were transferred into a ROS bag format through data streaming. Hence, the movements of CURI were transformed into the cobot coordinates by mapping between the marker and the end effector. The execution of whole-body motion was based on the reference trajectories and the corresponding impedance profiles of the end effector learned offline from demonstration via TP-LQT method. The human partner was wearing the EMG sensors during the experiments to monitor the muscle co-contraction, which was used to calculate the regulatory parameters of the cobot based on the LSTM-based regulation policy. To control expectancy and effort biases, each subject completed 25 randomized trials comprising five impedance control methods (five trials per method), with method identity masked and trial order fully randomized across different subjects.

The snapshots of the human-robot collaborative object transportation task with HI-ImpRSL method were given in Fig. \ref{Fig.exp_overall} with two subjects. The cobot and the human partners successfully lifted the object from the initial position, translated the object, and lowered the object to different target positions. The movement of the human-robot collaboration in these three stages was similar to the human-human demonstration. We also performed four comparative experiments with the same reference trajectory while the impedance variations of the end effector were different. The FIC method set $K_d$ = 300 $N/m$ and $D_d$ = 34.64 $N \cdot s/{m}$ in all three directions. The EMG-based method integrates biosignals to adapt impedance parameters based on task-specific rules, which was defined as a positive linear correlation in the leader-follower mode. The NN-based VIC method applied the interaction force-based stiffness estimation and used the variable impedance skill framework, which was learned from demonstrations based on a neural network. The HI-ImpRSL w/o LSTM method directly used the impedance profiles which learned from human demonstrations without the LSTM module.

\begin{table*}[htbp]
\caption{Interactive Forces Statistics Comparison of Task I (Mean $\pm$ SD)}
\vspace{-2mm}
\centering
\scriptsize
\begin{tabular}{l|ccc|ccc}
\hline
\addlinespace[0.2em]
Method & \multicolumn{3}{c|}{Force Mean (\textit{N})} & \multicolumn{3}{c}{Smoothness ($10^{-3}$ \textit{$N / s^2$})} \\
 & X & Y & Z & X & Y & Z \\
\addlinespace[0.2em]
\hline
\addlinespace[0.2em]
FIC & 1.66 $\pm$ 0.27 & 1.10 $\pm$ 0.10 & 5.82 $\pm$ 0.82 & 2.19 $\pm$ 0.11 & 2.23 $\pm$ 0.13 & 2.30 $\pm$ 0.25 \\
EMG-based VIC & 1.54 $\pm$ 0.29 & 1.11 $\pm$ 0.20 & 5.15 $\pm$ 0.92 & 2.17 $\pm$ 0.17 & 2.24 $\pm$ 0.16 & 2.62 $\pm$ 0.27 \\
NN-based VIC & 1.24 $\pm$ 0.32 & 0.99 $\pm$ 0.18 & 5.05 $\pm$ 0.57 & 2.01 $\pm$ 0.23 & 1.98 $\pm$ 0.17 & 2.27 $\pm$ 0.20 \\
HI-ImpRSL w/o LSTM & 1.34 $\pm$ 0.24 & 1.01 $\pm$ 0.12 & 5.53 $\pm$ 0.84 & 2.25 $\pm$ 0.30 & 2.22 $\pm$ 0.31 & 2.36 $\pm$ 0.17 \\
HI-ImpRSL & \textbf{1.18} $\pm$ 0.27 & \textbf{0.92} $\pm$ 0.12 & 5.27 $\pm$ 0.27 & \textbf{1.94} $\pm$ 0.20 & \textbf{1.86} $\pm$ 0.23 & \textbf{2.06} $\pm$ 0.17 \\
\addlinespace[0.2em]
\hline
\end{tabular}
\vspace{1mm}

\footnotesize
*Force Mean: the time-averaged absolute force magnitude; Smoothness: weighted RMSD of force jerk (second derivative); \\
*Values are aggregated over subjects S2–S5 as mean $\pm$ standard deviation.
\label{tab:forces_summary_compact}
\vspace{-4mm}
\end{table*}

We have recorded the interaction force from the cobot-side force/torque sensor to further verify the advantage of our method during the experiments. Unless otherwise specified, the reported “mean interaction force” is computed as the time-averaged absolute force magnitude. The results between CURI and subjects were shown in Table \ref{tab:forces_summary_compact}.

The subject-averaged results were given among S2-S5, while HI-ImpRSL delivers the best overall performance in both interaction force magnitude and smoothness. For force magnitude, HI-ImpRSL attains the lowest values in the $X$ and $Y$ directions ($1.18 \pm 0.27 N$ and $0.92 \pm 0.12 N$, respectively), outperforming FIC, EMG-based VIC, and the ablated variant HI-ImpRSL w/o LSTM ($1.34 \pm 0.24 N$ and $1.01 \pm 0.12 N$, respectively). In the $Z$ direction, NN-based VIC achieves the lowest mean force ($5.05 \pm 0.57 N$), while HI-ImpRSL remains competitive ($5.27 \pm 0.27 N$).

In terms of smoothness, quantified as the weighted RMSD of the second derivative of the interaction forces, HI-ImpRSL consistently achieves the smallest values across all axes ($1.94 \pm 0.20 \cdot 10^{-3} N/s^2$, $1.86 \pm 0.23 \cdot 10^{-3} N/s^2$, and $2.06 \pm 0.17 \cdot 10^{-3} N/s^2$ in $X$, $Y$, and $Z$ direction). These results indicate reduced high-frequency fluctuations and improved interaction stability compared with FIC, EMG-based VIC, NN-based VIC ($2.01 \pm 0.23 \cdot 10^{-3} N/s^2$, $1.98 \pm 0.17 \cdot 10^{-3} N/s^2$, and $2.27 \pm 0.20 \cdot 10^{-3} N/s^2$ in $X$, $Y$, and $Z$ direction), and HI-ImpRSL w/o LSTM. Overall, the aggregated statistics suggest that HI-ImpRSL most effectively suppresses interaction disturbances while maintaining the smoothest force profiles among the compared methods.

\subsection{Tai Chi Pushing Hands}
\subsubsection{Task Definition}

The second task is Tai Chi pushing hands, which is widely known as a popular Chinese kung fu because of its health benefits, especially for elders. As the basis of the pushing hands, the fixed step involves pushing motions in flat circles, vertical circles, and `8'-shaped circles \cite{wang2016biomechanical}. We select unimanual pushing of flat circles and bimanual pushing of vertical circles for collaboration (Fig. \ref{Fig.demonstration_transportation}).

Tai Chi pushing hands inherently features mutual adaptation, with partners alternating between stiff and compliance across the motion cycle. One leads with a stiffer push while the other yields compliantly, and the roles reverse on the return. Framed in rehabilitation terms, this alternating pattern aligns with passive–active training, where therapist-like guidance is paired with patient-like yielding. Hence, the cobot has to generate regulatory compliance behavior compared to the human partner in each stage of Tai Chi, which will be learned from the impedance regulation module.

\begin{table*}[t]
\caption{Interactive Forces Statistics Comparison of Task II and III (Mean $\pm$ SD)}
\vspace{-2mm}
\centering
\scriptsize
\begin{tabular}{l|c|ccc|ccc}
\hline
Task & Method & \multicolumn{3}{c|}{Force Mean (N)} & \multicolumn{3}{c}{Smoothness ($10^{-3}$ N/s$^{2}$)} \\
& & X & Y & Z & X & Y & Z \\
\hline
\addlinespace[0.2em]
\multirow{4}{*}{\textbf{Tai Chi Uni}}  
& FIC & 20.47 $\pm$ 7.40 & 6.00 $\pm$ 2.10 & 6.77 $\pm$ 3.12 & 5.40 $\pm$ 1.02 & 4.87 $\pm$ 0.82 & 5.52 $\pm$ 1.15 \\
& EMG-based VIC & 15.46 $\pm$ 6.60 & 5.78 $\pm$ 2.04 & 4.97 $\pm$ 1.40 & 5.15 $\pm$ 1.11 & 4.76 $\pm$ 0.59 & 3.95 $\pm$ 0.62 \\
& NN-based VIC & 22.54 $\pm$ 9.34 & 4.98 $\pm$ 1.87 & 4.25 $\pm$ 1.07 & \textbf{4.98} $\pm$ 0.96 & 4.14 $\pm$ 0.52 & 2.88 $\pm$ 0.61 \\
& HI-ImpRSL w/o LSTM & 18.76 $\pm$ 6.47 & 5.51 $\pm$ 2.14 & 4.14 $\pm$ 0.83 & 5.11 $\pm$ 0.89 & 5.12 $\pm$ 1.62 & 4.56 $\pm$ 1.04 \\
& HI-ImpRSL & 19.97 $\pm$ 8.34 & \textbf{4.66} $\pm$ 1.53 & \textbf{3.98} $\pm$ 1.24 & 5.26 $\pm$ 1.28 & \textbf{3.61} $\pm$ 0.36 & \textbf{2.45} $\pm$ 0.16 \\
\addlinespace[0.2em]
\hline
\hline
\addlinespace[0.2em]
\multirow{4}{*}{\textbf{Tai Chi Bi: Right}}
&FIC & 4.27 $\pm$ 0.98 & 4.82 $\pm$ 2.14 & 17.48 $\pm$ 3.35 & 3.93 $\pm$ 0.40 & 4.46 $\pm$ 0.27 & 4.97 $\pm$ 1.72 \\
&EMG-based VIC & 4.43 $\pm$ 1.66 & 4.62 $\pm$ 2.36 & 16.65 $\pm$ 4.05 & 3.82 $\pm$ 0.49 & 4.34 $\pm$ 0.47 & 5.45 $\pm$ 1.79 \\
& HI-ImpRSL w/o LSTM & 3.08 $\pm$ 0.73 & 4.20 $\pm$ 2.00 & 16.84 $\pm$ 3.10 & 3.46 $\pm$ 0.17 & 4.66 $\pm$ 1.48 & 4.98 $\pm$ 1.41 \\
&HI-ImpRSL & \textbf{2.70} $\pm$ 0.56 & \textbf{4.16} $\pm$ 1.80 & 16.09 $\pm$ 3.14 & \textbf{2.97} $\pm$ 0.26 & \textbf{3.68} $\pm$ 0.37 & \textbf{4.89} $\pm$ 2.41 \\
\addlinespace[0.2em]
\hline
\addlinespace[0.2em]
\multirow{4}{*}{\textbf{Tai Chi Bi: Left}}
&FIC & 2.95 $\pm$ 1.01 & 4.55 $\pm$ 1.90 & 13.07 $\pm$ 5.78 & 4.04 $\pm$ 0.46 & 4.65 $\pm$ 1.10 & 6.19 $\pm$ 2.50 \\
&EMG-based VIC & 2.96 $\pm$ 0.36 & 3.95 $\pm$ 0.43 & 12.35 $\pm$ 2.57 & 4.10 $\pm$ 0.35 & 4.49 $\pm$ 1.24 & 5.36 $\pm$ 2.05 \\
&HI-ImpRSL w/o LSTM & 3.17$\pm$ 1.24 & 3.48 $\pm$ 0.83 & 13.07 $\pm$ 2.82 & 3.76 $\pm$ 0.33 & 3.93 $\pm$ 0.50 & 5.67 $\pm$ 1.69 \\
&HI-ImpRSL & \textbf{2.28} $\pm$ 0.69 & \textbf{3.08} $\pm$ 0.91 & 13.32 $\pm$ 4.78 & \textbf{2.65} $\pm$ 0.34 & \textbf{3.33} $\pm$ 0.53 & \textbf{4.82} $\pm$ 1.92 \\
\addlinespace[0.2em]
\hline
\addlinespace[0.2em]
\multirow{4}{*}{\textbf{Sawing}}
&FIC & 5.51 $\pm$ 6.71 & 3.21 $\pm$ 1.37 & 6.80 $\pm$ 3.86 & 8.45 $\pm$ 2.18 & 5.07 $\pm$ 0.97 & 10.26 $\pm$ 6.19\\
&EMG-based VIC & 5.74 $\pm$ 6.30 & 2.95 $\pm$ 1.67 & 7.13 $\pm$ 3.65 & 6.72 $\pm$ 3.82 & 4.33 $\pm$ 0.89 & 9.13 $\pm$ 5.64 \\
&HI-ImpRSL w/o LSTM & 8.18$\pm$ 8.06 & 2.53 $\pm$ 1.75 & 7.67 $\pm$ 4.08 & 5.88 $\pm$ 3.21 & 3.68 $\pm$ 0.57 & 8.05 $\pm$ 3.88 \\
&HI-ImpRSL & 7.71 $\pm$ 8.13 & \textbf{1.62} $\pm$ 0.96 & \underline{9.18} $\pm$ 2.98 & 6.32 $\pm$ 4.38 & \textbf{3.12} $\pm$ 0.34 & \textbf{7.47} $\pm$ 3.52 \\
\addlinespace[0.2em]
\hline
\end{tabular}
\label{tab:taichi_forces_summary}
\vspace{-4mm}
\end{table*}

\subsubsection{Trajectory and Impedance Generation from Demonstration}
The data collection setup of confrontational Tai Chi pushing hands was the same as the collaborative transportation task. Two sub-tasks including the unimanual flat circle (Fig. \ref{Fig.demonstration_transportation}.II) and bimanual vertical circle (Fig. \ref{Fig.demonstration_transportation}.III) pushing hands were performed by two human demonstrators. Since we only calibrated one upper arm of each demonstrator, the parameters of the other arm were set to the same values as the calibrated one for simplification in the bimanual task. 

The motion representation and generation results of the unimanual Tai Chi pushing tasks are shown in Fig. \ref{Fig.tpgmm_boxcarrying}.II. Besides, impedance regulation policies of both Tai Chi tasks were learned by the LSTM networks and one example of training and testing predictions was given in Fig. \ref{Fig.lstm_box}.II. The trained models were further applied to get the regulatory factors for the cobot to tune compliance.

\subsubsection{Experimental Results}
In the real-world experiments (as illustrated in Fig. \ref{Fig.exp_overall}), the setup of the cobot and the human partner was almost the same as the first task, except that two robotic arms were applied in the second Tai Chi task. Four impedance regulation methods for the cobot were performed with every subject in random order, which included 5 trials for each method and a total of 20 trials for each subject. 

The snapshots of human-robot confrontational pushing hands with unimanual flat circle pushing and bimanual vertical circle pushing were shown in Fig. \ref{Fig.exp_overall}. In each motion cycle, CURI utilized the learned regulation policy and achieved the task. For instance, during the unimanual arm pushing task, CURI initially exerted high impedance to push the human towards one side and subsequently reduced the pushing force in the latter half of the movement. Specifically, CURI right arm maintained high stiffness and damping in the initial phase and decreased the impedance in the latter half. 

The statistical analysis of interactive forces with four subjects is given in Table \ref{tab:taichi_forces_summary}, which illustrates the average performance of S2-S5 in the form of MEAN $\pm$ SD. Our proposed HI-ImpRSL exhibits the most favorable overall performance in both interaction force magnitude and smoothness across the collaborative Tai Chi tasks. For the unimanual task, where the motion primarily lies in the $XY$ plane and the $Z$-axis force best reflects disturbance rejection, HI-ImpRSL attains the lowest $Z$-direction mean force ($3.98 \pm 1.24 N$), outperforming FIC ($6.77\pm3.12N$), EMG-based VIC ($4.97\pm1.40N$), NN-based VIC ($4.25\pm1.07N$), and HI-ImpRSL w/o LSTM ($4.14\pm0.83N$). It also achieves the smallest $Y$-direction force ($4.66\pm1.53N$), indicating improved suppression of out-of-plane disturbances and higher cooperativity.

For bimanual tasks, where the trajectory is mainly in the $YZ$ plane and the force on the $X$-axis is most indicative, HI-ImpRSL achieves the lowest mean forces in the $X$-direction on both arms. On the right arm, it records $2.70\pm0.56N$, outperforming FIC ($4.27\pm0.98N$), EMG-based VIC ($4.43\pm1.66N$), and HI-ImpRSL w/o LSTM ($3.08\pm0.73N$). On the left arm, it achieves $2.28\pm0.69N$, lower than FIC ($2.95\pm1.01N$), EMG-based VIC ($2.96\pm0.36N$), and HI-ImpRSL w/o LSTM ($3.17\pm1.24N$). These reductions corroborate HI-ImpRSL’s effectiveness in minimizing interaction forces and enhancing bilateral coordination.

Smoothness, quantified as the weighted RMSD of the second derivatives of the force signals, further underscores the stability of HI-ImpRSL. In the unimanual task, it achieves the lowest values along $Y$ and $Z$ ($3.61\pm0.36\cdot10^{-3}N/s^2$ and $2.45\pm0.16\cdot10^{-3}N/s^2$, respectively), outperforming all baselines. Besides, NN-based VIC attains the smallest value on $X$ ($4.98\pm0.96\cdot10^{-3}N/s^2$) while HI-ImpRSL remains competitive ($5.26\pm1.28\cdot10^{-3}N/s^2$). For bimanual tasks, HI-ImpRSL consistently yields the lowest smoothness metrics on all axes on both the right arm ($2.97\pm0.26,3.68\pm0.37,4.89\pm2.41\cdot10^{-3}N/s^2$) and the left arm ($2.65\pm0.34,3.33\pm0.53,4.82\pm1.92\cdot10^{-3}N/s^2$). Overall, the aggregated statistics indicate that HI-ImpRSL most effectively suppresses interaction disturbances while maintaining the smoothest force profiles across the evaluated Tai Chi tasks.

\subsection{Collaborative Sawing}
We further introduce a collaborative sawing task to evaluate mutual adaptation under tool-environment dynamics. A human and the cobot jointly operate a hand saw to cut a wooden block (as shown in Fig. \ref{Fig.exp_overall}). We adopt a hybrid position-force scheme that decouples planar stroke generation from normal-force regulation. In the $XY$ plane, a reference trajectory is learned via TP-LQT from demonstrations and adapted online to the current frame. The cobot’s in-plane stiffness and damping are modulated by the partner’s muscle activation, using processed EMG envelopes to adjust impedance gains for phase-aware compliance and stability under varying friction. Along the $Z$ axis, a force controller maintains a constant normal force (10$N$) to ensure cutting engagement while compensating surface irregularities. Interaction forces are measured at the end effector, and EMG provides a proxy of human effort for impedance regulation.

Quantitatively, HI-ImpRSL attains the smallest off-axis interaction forces in the $Y$ direction ($1.62 \pm 0.96N$) compared to the other three methods. Regarding the smoothness, our method consistently achieves smoother interaction ($3.12 \pm 0.34 \cdot 10^{-3}N/s^{2}$). Along $Z$ direction, where the controller regulates a 10 $N$ normal force, HI-ImpRSL exhibits $9.18 \pm 2.98N$ (underlined as closest to the 10 $N$ reference), demonstrating superior force-tracking accuracy compared with FIC ($6.80 \pm 3.86N$), EMG-based VIC $(7.13 \pm 3.65N$), and the ablation ($7.67 \pm 4.08N$). These improvements confirm that HI-ImpRSL simultaneously enhances planar compliance and normal-force fidelity in the collaborative sawing task.

\begin{figure}[t]
\centering
\includegraphics[width=0.95\linewidth]{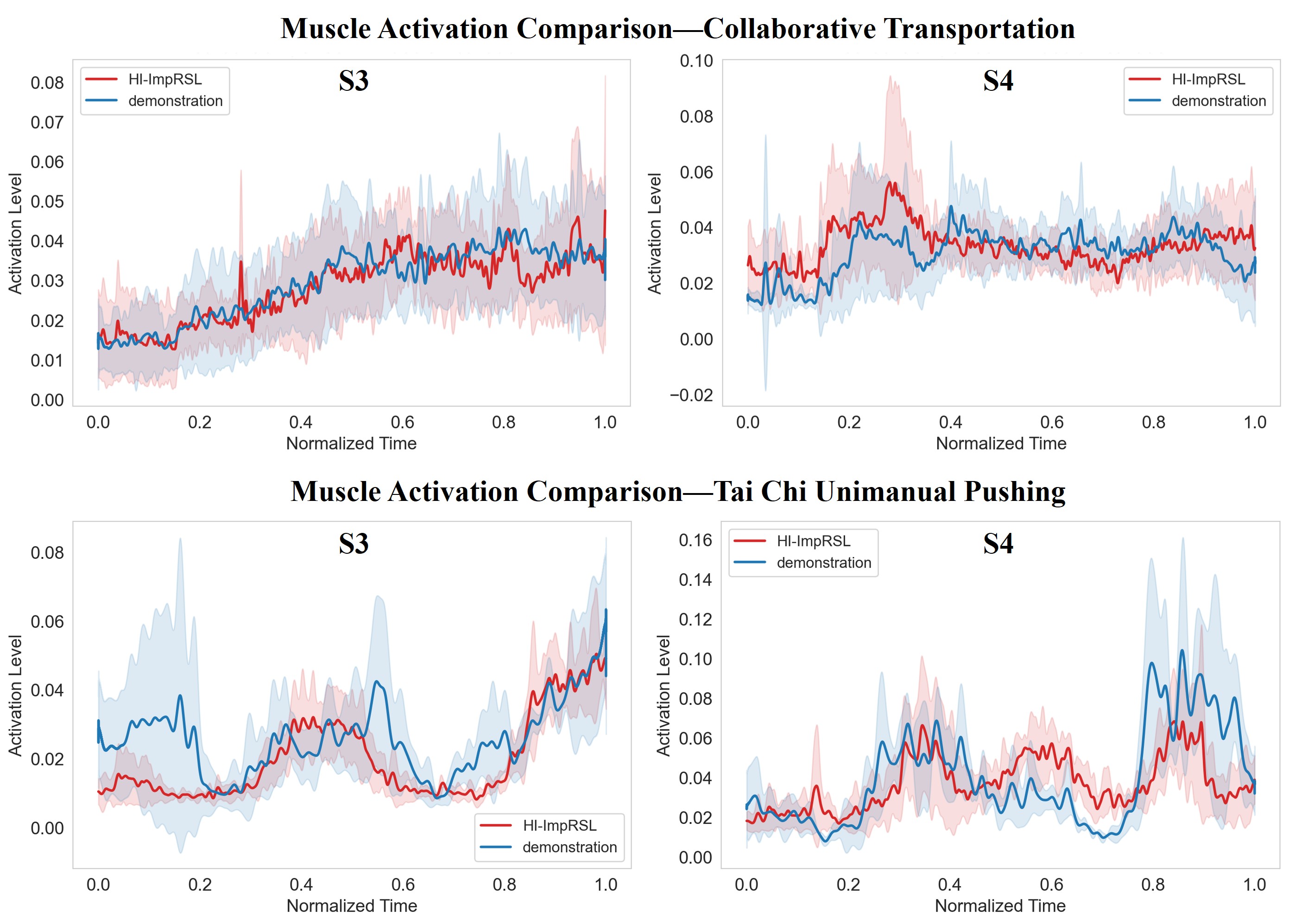} \vspace{-4mm}
\caption{The muscle activation comparison between HI-ImpRSL and demonstration, including collaborative transportation (top) and Tai Chi unimanual pushing (bottom) performed by S3 and S4.} 
\label{Fig.muscle_comparison} \vspace{-3mm}
\end{figure}

\subsection{Discussion}
To compare the collaboration dynamics of the leader-follower and mutual adaptation modes, we analyzed the smoothness and variations of the interaction forces recorded during human-robot collaboration. In the transportation task, interaction forces were smoother and exhibited lower variations in the main direction of movement. In contrast, the Tai Chi pushing hands task showed greater variability due to its mutual adaptation mode, highlighting the task’s bidirectional compliance regulation and dynamic role switching. These results demonstrate the contrasting interaction dynamics between the two types of tasks, with the leader-follower emphasizing stable compliance and the mutual adaptation requiring more adaptive and variable behaviors, validating the framework’s adaptability across different collaboration modes.

We further compared the muscle activation trends of S3 and S4 during the demonstrations and HRC with our HI-ImpRSL framework across the collaborative transportation task and the Tai Chi unimanual pushing task (Fig. \ref{Fig.muscle_comparison}). The results revealed that the overall trends in muscle activation were highly similar, further validating that our proposed method can effectively generate human-like compliance behaviors. However, we observed a notable difference in the initial phase of the Tai Chi task for S3, where the muscle activation levels were less aligned. This discrepancy is attributed to the tendency to exert force during the preparatory phase when collaborating with a human partner, whereas with the robotic system, force application occurred only after the motion had begun. Aside from this, the subsequent muscle activation trends were consistent.

It is worth noting that beyond Tai Chi pushing hands, our proposed HI-ImpRSL can be extended to rehabilitation or medical robotics by using human-human therapeutic demonstrations to derive task-specific trajectories and impedance profiles. The LSTM-based regulation can adapt assistance or resistance to patient effort with safety-aware compliance, indicating broad clinical applicability.

\section{Conclusions}\label{VI} 

In this paper, we propose a novel HI-ImpRSL framework to acquire impedance regulation skills from human-human collaboration. By calibrating the human dynamic model, we estimate endpoint stiffness and then compute reference trajectories based on the interaction model with these measured data. A TP-LQT method is proposed to represent human movements and generate a new reference trajectory for the cobot under new situations. Besides, the impedance regulation skills of each task are learned through the LSTM networks to construct the regulation modules. Finally, we apply a human-like cobot CURI to HRC tasks to perform object transportation and Tai Chi pushing hands with a human partner. We propose a whole-body impedance controller for this unique cobot and put forward a variable impedance control method based on the human-human demonstrations. The real-world platform we built successfully achieved the collaborative object transportation task, Tai Chi pushing hands tasks, and collaborative sawing tasks, demonstrating effective adaptive impedance regulation. This study bridges human cognitive behaviors and robotic skill learning, advancing HRC and providing a foundation for future research in robotics.

Nevertheless, we limited our analysis to only the arms and torso of our cobot during collaboration, which constrains the cobot workspace and functional capacity to some extent. Additionally, considerations of human safety and physical ergonomics were not addressed by the cobot. To enhance task execution, our future work will incorporate a capture system that includes feedback on human kinematics and dynamics. Moreover, the current setup for human-human demonstrations is complex, and the transferability of learned behaviors from one task to another remains underdeveloped. Future research will focus on designing more robust, generalized data representations inspired by cognitive systems, allowing the framework to extract richer interaction patterns and improve its adaptability to diverse collaborative scenarios.

\bibliographystyle{ieeetr}
\bibliography{ref_hrc}

\vspace{11pt}
\end{document}